\pdfoutput=1
\documentclass[11pt]{article}
\usepackage{fancyvrb}
\usepackage{ACL2023}

\usepackage{tcolorbox}

\usepackage[frozencache=true,cachedir=minted-cache]{minted}

\definecolor{emerald}{rgb}{0.31, 0.78, 0.47}
\definecolor{babyblue}{rgb}{0.54, 0.81, 0.94}
\definecolor{bananayellow}{rgb}{1.0, 0.88, 0.21}

\usepackage{amsmath}

\usepackage{times}
\usepackage{latexsym}

\usepackage[T1]{fontenc}
\newcommand{\ignore}[1]{}

\usepackage[utf8]{inputenc}

\usepackage{microtype}

\usepackage{inconsolata}

\usepackage{xcolor}
\usepackage{listings}
\usepackage[textsize=footnotesize]{todonotes}

\usepackage{amssymb}% http://ctan.org/pkg/amssymb
\usepackage{pifont}% http://ctan.org/pkg/pifont
\newcommand{\cmark}{\ding{51}}%
\newcommand{\xmark}{\ding{55}}%

\newenvironment{itemizesquish}[2]{\begin{list}{\labelitemi}{\setlength{\itemsep}{#1}\setlength{\labelwidth}{#2}\setlength{\leftmargin}{\labelwidth}\addtolength{\leftmargin}{\labelsep}}}{\end{list}}

\title{The Larger They Are, the Harder They Fail: \\ Language Models do not Recognize Identifier Swaps in Python}
\author{$\,\,\,\,\,\,\,\,\,\,\,\,\,\,\,\,\,\,\,\,\,\,\,\,$Antonio Valerio Miceli-Barone$^1$\Thanks{$\,\,$Equal contribution.} $\qquad$ \\ \texttt{amiceli@ed.ac.uk} 
\And
Fazl Barez$^{1*}$ \\ \texttt{f.barez@ed.ac.uk}
\AND
Ioannis Konstas$^2$ \\ \texttt{i.konstas@hw.ac.uk}
\And
Shay B. Cohen$^1$ \\ \texttt{scohen@inf.ed.ac.uk}
\AND
\textnormal{\normalsize $^1$ School of Informatics, University of Edinburgh}
\\
\normalsize $^2$ School of Mathematical and Computer Sciences, Heriot-Watt University
}

\begin{document}
\maketitle
\begin{abstract}
Large Language Models (LLMs) have successfully been applied to code generation tasks, raising the question of how well these models understand programming.
Typical programming languages have invariances and equivariances in their semantics that human programmers intuitively understand and exploit, such as the (near) invariance to the renaming of identifiers. We show that LLMs not only fail to properly generate correct Python code when default function names are swapped, but some of them even become more confident in their incorrect predictions as the model size increases, an instance of the recently discovered phenomenon of \textit{Inverse Scaling}, which runs contrary to the commonly observed trend of increasing prediction quality with increasing model size. Our findings indicate that, despite their astonishing typical-case performance, LLMs still lack a deep, abstract understanding of the content they manipulate, making them unsuitable for tasks that statistically deviate from their training data, and that mere scaling is not enough to achieve such capability.

\end{abstract}

\section{Introduction}
\label{SEC:INTRO}
Pretrained Large Language Models (LLMs) are rapidly becoming one of the dominant paradigm for large variety of language tasks \cite{LLM_few_shot_learners, PALM}, including programming code generation and completion \citep{openai2021codex, Li2022AlphaCode}.
LLMs have demonstrated increasing performance with increasing model size\footnote{%\shaycomment{sorry, this is very opaque - do you mean they are fixed in *our* experiments? in a certain type of models? what do you mean by ``empirical law'', what is it? $\rightarrow$} {\color{red} I mean the optimal size-data-compute laws found in the scaling papers. - Antonio}
Since model capacity in number of parameters and pretraining dataset size are balanced according to a design law that is fixed for each model family and is intended to empirically maximize the pretraining set likelihood given a compute budget \citep{Scaling_laws_LM}, for the reminder of this paper we will jointly refer to them as "model size".} on many practical tasks \cite{Scaling_laws_LM, Scaling_laws_transfer} including programming tasks \cite{Nijkamp2022codegen}, recently, however, researchers have identified a number of tasks that exhibit \textit{inverse scaling}, where output quality decreases, rather than increase, with increasing model size.

% \begin{figure}
%     \centering
%     \includegraphics[width=0.45\textwidth]{drake_meme_bultins_swap_cropped.pdf}
%     \caption{Given a Python prompt which swaps of two builtin functions, large language models prefer the incorrect but statistically common continuation to the correct but unusual one.}
%     \label{fig:intro:meme_example}
% \end{figure}

\begin{figure}
    \centering
    \setlength\tabcolsep{1.5pt} % default value: 6pt
    \begin{tcolorbox}
    \texttt{len, print = print, len} \\
    \texttt{\textcolor{blue}{def} \textcolor{brown}{print\_len}(x)}:\\
    \texttt{$\, \,$\textcolor{brown}{\-\ \-\ ``Print the length of x''}}
    \end{tcolorbox}
    \begin{tabular}{cc}
    \vspace{-0.15in} \hspace{0.21\textwidth} &     \hspace{0.215\textwidth} \\
    \textcolor{green}{\cmark} \texttt{\textcolor{brown}{len}(\textcolor{brown}{print}(x))} & 
    \textcolor{red}{\xmark}  \texttt{\textcolor{brown}{print}(\textcolor{brown}{len}(x))}  \\
& \fbox{\textbf{LLM preference}}
    \end{tabular}
    
    \caption{Given a Python prompt (on top) which swaps of two builtin functions, large language models prefer the incorrect but statistically  common continuation (right) to the correct but unusual one (left).
    %\shaycomment{mmm, I wonder if we should switch between the X and the V so that it indicates what is "truly" correct", and the picture above indicates the LLM's preference}
    %{\color{red} Yes, also the docstring should be indented. - Antonio}
    }
    \label{fig:intro:meme_example}
\end{figure}
%\shaycomment{May I suggest to use this version of Figure 1, as I find the previous version to waste a lot of space?}

\begin{figure*}[t]
    \centering
    \includegraphics[width=0.9\textwidth]{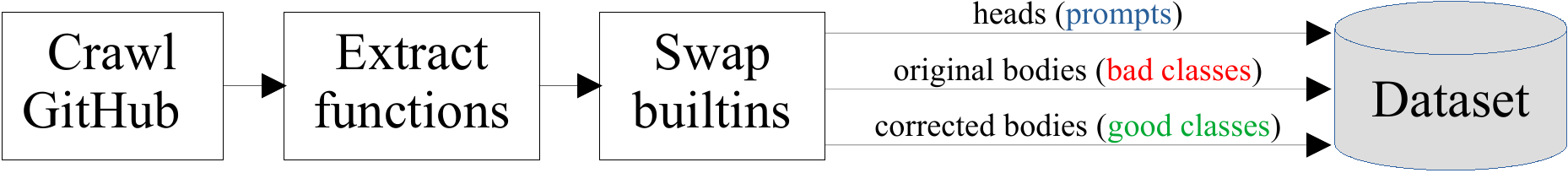}
    \caption{Data generation pipeline (see Appendix~\ref{sec:appendix:datasetexample} for an example):
    1. Crawl repositories from GitHub, filtered by language, license, stars, and size. 2. Extract top-level functions with docstrings and references to at least two callable builtins 
    %\shaycomment{are you sure it is not built-in that you need to use in most caes rather than builtin?}. 
    3. For each function, choose two builtins to swap and generate: a) header with builtin swap statement, function declaration with decorators, docstring b) original function body, c) corrected body with the builtins swapped consistently with the swap statement. 4. Store as a binary classification task: a) head = classifier input, b) original body = bad class, c) corrected body = good class.}
    \label{fig:method:datageneration}
\end{figure*}

Tasks with inverse scaling generally either involve social biases \cite{ParrishBias2021,LLM_x_risk_extrap}, where the larger models (arguably correctly) learn undesirable biases from biased training sets, or involve examples of natural language that are highly atypical but still easily understandable by a human \cite{mckenzie2022round1}.
These tasks may involve unusual discourse pragmatics or they may require reasoning about counterfactual knowledge, however, since they tend to be highly artificial, it could perhaps be argued that they are edge cases which may not represent serious failure modes for practical applications.
In this paper we present a novel type of inverse scaling task involving Python code generation under a redefinition of default identifiers. %\cite{social_bias_LLM}
This has both practical implications (redefinition of default identifiers is a meta-programming technique used in popular libraries), and broader scientific implications, as it shows that LLMs fail to reason about the deep, abstract semantic structure of programming languages, and these flaws are not ameliorated, but in fact may be even worsened, by increasing model size.

Programming languages have precise and  well-defined syntax and semantics which makes them especially suited to automatic analysis and procedural generation.
They are scientifically interesting because they can be used for automatic generation of examples of coding problems and their evaluation against an objective ground truth, whereas most NLP tasks have enough ambiguity that require human annotation in order to produce high-quality examples. 
%hence programming languages are convenient domains to study the learning capabilities of LLMs beyond simple lexical generalization.
Furthermore, this research is also of practical importance for software engineering tools that use LLMs, such as GitHub Copilot,\footnote{\url{https://github.com/features/copilot}} which are starting to be widely adopted by developers.

%In addition, we discuss why this type of task is important for testing the performance of natural language models, and present evidence of a bounded level of performance exhibited by current state-of-the-art models. We also describe how our approach to inverse scaling can be extended to other programming languages, given their syntactic and semantic structures. Finally, we make recommendations for future work aimed at resolving the issues uncovered by our experiments.

%These instructions are for authors submitting papers to ACL 2023 using \LaTeX. They are not self-contained. All authors must follow the general instructions for *ACL proceedings,\footnote{\url{http://acl-org.github.io/ACLPUB/formatting.html}} as well as guidelines set forth in the ACL 2023 call for papers.\footnote{\url{https://2023.aclweb.org/calls/main_conference/}} This document contains additional instructions for the \LaTeX{} style files.
%The templates include the \LaTeX{} source of this document (\texttt{acl2023.tex}),
%the \LaTeX{} style file used to format it (\texttt{acl2023.sty}),
%an ACL bibliography style (\texttt{acl\_natbib.bst}),
%an example bibliography (\texttt{custom.bib}),
%and the bibliography for the ACL Anthology (\texttt{anthology.bib}).

\section{Methodology}
\label{SEC:METHOD}

We describe the motivation behind our task (\S\ref{SEC:METHOD:TASK_MOTIVATION})
and the task itself (\S\ref{SEC:METHOD:TASK_DESCRIPTION}), followed by the way
we collected the data for the task (\S\ref{SEC:METHOD:DATA_GENERATION}).

We release our dataset as well as the code used to generate it and replicate our experiments\footnote{\url{https://github.com/Avmb/inverse_scaling_prize_code_identifier_swap.git}}.

\subsection{Task Motivation}
\label{SEC:METHOD:TASK_MOTIVATION}

Turing-complete languages have invariances and equivariances, %\footnote{Informally, a transformation $T$ (e.g. a syntactic substitution) is an invariance for a function $f$ (e.g. the denotational semantics of a programming language) if $\forall{x}, f(x)=f(T(x))$, or it is an equivariance if $\forall{x}, T(f(x))=f(T(x))$.} 
%\shaycomment{I sugget you explain here what the difference is between the two} 
making it possible to express the same function by multiple programs (see Appendix \ref{sec:appendix:equivinv} for formal definitions). While determining semantic equivalence is undecidable in the general case \cite{Rice1953ClassesOR}, sometimes it can be determined by pure syntactic analysis.
For instance, $\alpha$-equivalence, invariance under the consistent renaming
of identifiers such as variable or function names, can be decided using syntactic analysis.
%\shaycomment{what does alpha-equivalence mean? and should it be $\alpha$-equivalence?} 
%refers to the property of many common programming languages of being invariant with respect to a renaming of identifiers such as variable or function names.

\ignore{It can be argued that a proper scientific understanding of a system consists of identifying its invariances and equivariances which define laws that apply even out of the distribution of the observed data.
Identifying such relations from empirical evidence is, broadly speaking, the process of scientific discovery in disciplines such as physics or chemistry.\todo{is there anything to cite here? Fazl: yes
}}

Proper understanding of the semantics of a programming language requires identifying its invariances and equivariances, as opposed to ``shortcut learning'' \citep{Geirhos2020Shortcut} which instead exploits many weak, spurious correlations that do not generalize out of the observed data distribution.
We propose a task %(\S\ref{SEC:METHOD:TASK_DESCRIPTION}) 
based on the approximate $\alpha$-equivalence of Python code,
%\footnote{Python does not have exact $\alpha$-equivalence due to the existence of reflection and  evaluation facilities that can access and evaluate identifier names as strings at runtime. In practice, however, most Python functions do not use such facilities.}
in order to evaluate how well LLMs master the semantics of Python.
%Our task switches between common identifiers in pieces of code, and tests whether LLMs are able to cope with that change and embed it into their responses when requested to generate code.

\ignore{Our task provides useful information about the limitations of LLMs when applied to code,
and indicates that software engineering tools such as Copilot have shortcoming when
it comes to understanding the semantics of programs, even in simple cases that are straightforward for humans to identify and resolve.}

%that involves the redefinition of common identifiers, which is usual but it is sometimes used as a meta-programming technique in various libraries, and therefore should be correctly handled by code analysis and generation tools.
%\shaycomment{$\leftarrow$ I think such motivation should also be briefly mentioned in the introduction, perhaps more generally.}

\subsection{Task Description}
\label{SEC:METHOD:TASK_DESCRIPTION}

We consider code snippets in Python 3. Python allows to redefine \emph{builtin} functions\footnote{Predefined functions that the language exposes to the user.} by reassigning their identifiers.
For instance, the statement 
%\shaycomment{center what's below?}
%\todo{anyone knows how to center verbatim?}
%\begin{lstlisting}[language=Python]
%\begin{verbatim}
\begin{center}
\begin{BVerbatim}
len, print = print, len
\end{BVerbatim}
\end{center}

\noindent swaps the identifiers for the builtin functions \texttt{len} and \texttt{print}.
Any function defined following that identifier swap would have to refer to the builtin function \texttt{len} by the identifier \texttt{print} and vice versa.

We consider a code generation task where the model is given a top-level function \emph{declaration}, followed by a \emph{docstring} (which typically describes the behavior of the function in natural language) and has to generate the rest of the body of the function, similar to \citet{miceli-barone-sennrich-2017-parallel}, but with the caveat that we prepend to %\shaycomment{the}
the declaration a \emph{statement} that swaps two Python builtin functions that are expected to be used in the function body.
Specifically, in line with the format of the Inverse Scaling Prize\footnote{\url{https://github.com/inverse-scaling/prize}} we define our \textbf{Builtin identifier swap} task as a binary classification task where the input of each example is the concatenation of a swap statement, function declaration (with optional decorators) and docstring. A ``bad'' output for such input is a function body that uses the builtin functions according to their usual meaning, ignoring the swap statement. In contrast, the ``good'' output is a function body where the builtin functions are used consistently with the swap statement. To assess the success of the model in distinguishing between the ``bad'' and the ``good'' output, we compute the likelihood of each output given the input provided as a prompt (Figure~\ref{fig:intro:meme_example}, Appendix~\ref{sec:appendix:datasetexample}).

\subsection{Data Collection}
\label{SEC:METHOD:DATA_GENERATION}

Similar to \citet{miceli-barone-sennrich-2017-parallel}, our dataset collection procedure involves scraping code from GitHub using the \texttt{PyCodeSuggest} library\footnote{\url{https://github.com/uclnlp/pycodesuggest}} \citep{pycodesuggest} to download Python repositories with at least 100 stars, of size at most 200 MB and which mention the use of the Open Source CC-BY-4.0 license\footnote{\url{https://creativecommons.org/licenses/by/4.0/}} in their README.
Our final dataset includes 559 repositories downloaded on 16 December 2022.
We then parse the \textbf{.py} files in each repository with the Python 3 \texttt{ast} module to make sure that they contain valid code. We extract 1,000 randomly chosen top-level functions that each contain a docstring and that reference at least two callable builtin identifiers, as defined by the \texttt{builtins} module.
For each of these extracted functions, we randomly choose %\shaycomment{$\leftarrow$ I suggest you be consistent about the tense you use for describing this process, and probably present in my opinion is better, but definitely not mixed.} 
two builtin functions and generate the corresponding swap statement, function declaration (with decorators) and docstring as the example prompt, the original function body (regenerated from the abstract syntax tree with the \texttt{astunparse} module\footnote{\url{https://pypi.org/project/astunparse/}}%, which removes comments and provides a consistent style for indentation, spacing, constants
) as the ``bad'' output and the function body where the two selected builtins are swapped consistently with the swap statement as the ``good'' output (Figure~\ref{fig:method:datageneration}).

 Note that functions can in principle access the builtin identifiers as strings using reflection and evaluation facilities, which may require a full static analysis of the code to identify and is undecidable in the general case.
 Since our method uses purely syntactic substitutions, there might be cases where the ``good'' %\shaycomment{what "good" means here?} 
outputs do not maintain the expected function behavior.
In practice, this dynamic access of identifiers at runtime is rare with builtin identifiers and therefore does not pose an issue.

%, extracting the top-level functions with contain a docstring, and then randomly selecting two of the builtin functions from each function call. An associated prompt is generated from the function declarations up to the docstring, and two classes are produced - one with the correct substitution and one without - for the remaining portion after the docstring.

%In this study, we explored the utility of a novel dataset generation procedure for automatic code completion by scraping Python code from GitHub and extracting the top-level functions with a docstring. We randomly selected two of the builtin functions from each function call and associated prompt from the function declarations up to the docstring. We also evaluated two classes - one with the correct substitution and one without - for the remaining portion after the docstring. 

\section{Experiments}
\label{SEC:EXPERIMENTS}

We next describe our experiments with a likelihood calculation of correct and incorrect completions (\S\ref{SEC:EXPERIMENTS:LOSS}) and chat LLMs (\S\ref{SEC:EXPERIMENTS:CHATLLMS}),
and then present a qualitative analysis (\S\ref{SEC:EXPERIMENTS:QUALITATIVE}).

\paragraph{Computational resources} We spent approximately 130 US dollars, including donated credits, to use the OpenAI LLMs through their publicly accessible API.

We also used a small amount of machine-hours on the Baskerville Tier 2 HPC platform\footnote{\url{https://www.baskerville.ac.uk/}} equipped with NVIDIA A100 GPUs.
While this is a high-end system, our experiments on the open source models can be also practically run on consumer-grade machines with gaming GPUs.

\subsection{Completion Likelihood}
\label{SEC:EXPERIMENTS:LOSS}
For our main set of experiments, we evaluate our dataset on families of auto-regressive language models (OpenAI GPT-3, Salesforce CodeGen, Meta AI OPT) and one family of sequence-to-sequence conditional auto-regressive language models (Google FLAN-T5).
All  models are based on the Transformer architecture \citep{Vaswani2017Transformer} and pretrained on large datasets scraped from the Internet (full details in Appendix~\ref{SEC:EXPERIMENTS:MODELS}).
%for their description.

\begin{figure*}[ht]
\centering
\includegraphics[width=0.49\linewidth]{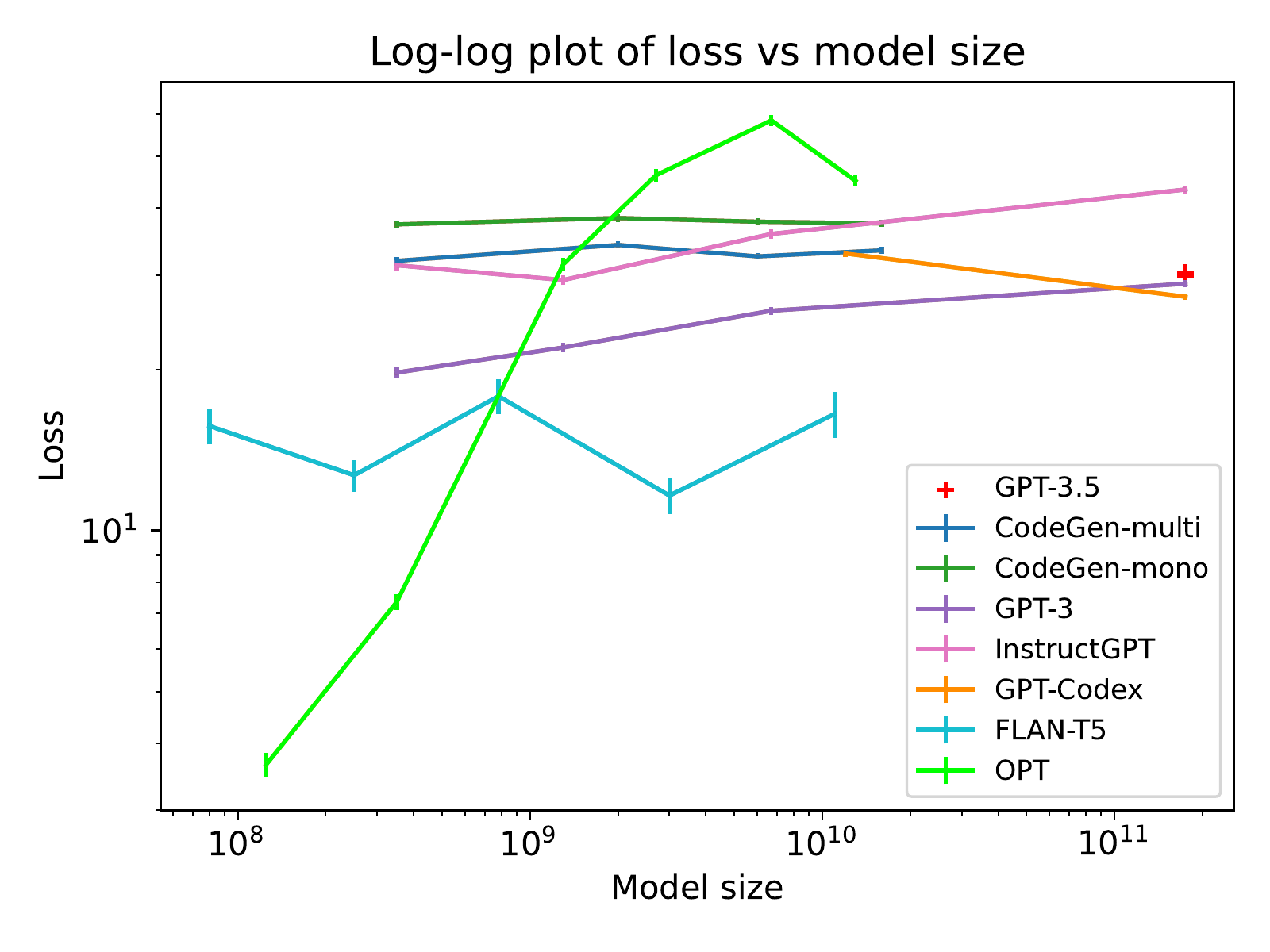}
\includegraphics[width=0.49\linewidth]{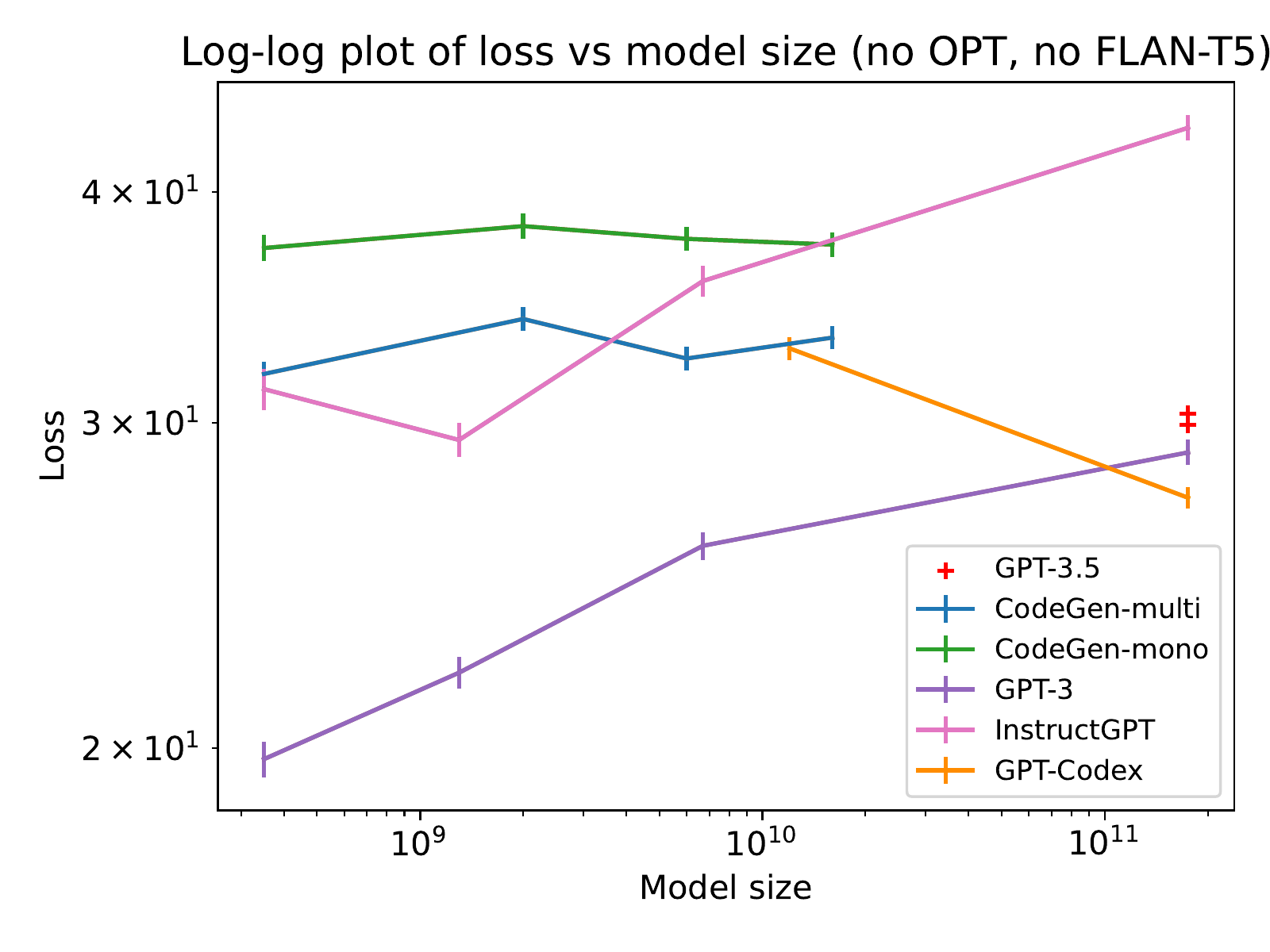}
\caption{Classification loss over model size. Left: all models. Right: all models except Meta AI OPT and Google FLAN-T5 families.}
 \label{fig:experiments:results:all}
\end{figure*}

%\subsection{Results}
%\label{SEC:EXPERIMENTS:RESULTS}
\paragraph{Results}
We evaluate our datasets on the models using a modified version of the Inverse Scaling Prize evaluation code.\footnote{Original:~\url{https://github.com/naimenz/inverse-scaling-eval-pipeline}, our version:~\url{https://github.com/Avmb/inverse-scaling-eval-pipeline.git}}
We report the results for all models in Figure~\ref{fig:experiments:results:all}. %, and we zoom the scale for the models excluding the OPT and FLAN-T5 families in Fig. \ref{fig:experiments:results:allnooptnoflant5}.
The graphs show the classification loss averaged over the examples for each model, with standard errors represented as error bars.

\begin{table}[h!]
\centering

\begin{tabular}{lrrr}
\hline
 Model family   & Pearson & Spearman  & Kendall         \\
\hline
 OPT            & \textbf{0.94} \ignore{ ($\mathbf{p}$=0.01)} & \textbf{0.83} \ignore{($\mathbf{p}$=0.04)} & \textbf{0.73} \ignore{($\mathbf{p}$=0.06) }\\
 GPT-3          & \textbf{0.97} \ignore{($\mathbf{p}$=0.03)} & \textbf{1.00}\ignore{ ($\mathbf{p}$=0.00)} & \textbf{1.00} \ignore{($\mathbf{p}$=0.08)} \\
 InstructGPT    & \textbf{0.94} \ignore{($\mathbf{p}$=0.06)} & 0.80 \ignore{($\mathbf{p}$=0.20)}          & 0.67 \ignore{($\mathbf{p}$=0.33)}          \\
 CodeGen-multi  & 0.46 & 0.40 \ignore{($\mathbf{p}$=0.60) }         & 0.33\ignore{ ($\mathbf{p}$=0.75) }         \\
 CodeGen-mono   & 0.10 \ignore{($\mathbf{p}$=0.90)}          & 0.20 \ignore{($\mathbf{p}$=0.80) }         & 0.00\ignore{ ($\mathbf{p}$=1.00)}          \\
 GPT-Codex      & -1.00\ignore{ ($\mathbf{p}$=1.00)  }       & -1.00 \ignore{($\mathbf{p}$=nan) }         & -1.00 \ignore{($\mathbf{p}$=1.00)}         \\
 FLAN-T5        & 0.01 \ignore{($\mathbf{p}$=0.99)}          & 0.10 \ignore{($\mathbf{p}$=0.87)   }       & 0.00\ignore{ ($\mathbf{p}$=1.00) }         \\
\hline
\end{tabular}
\caption{Correlation coefficients between log-model size and log-loss for each model family. Bolded values indicate inverse scaling at p-value < 0.1. The two text-based GPT-3.5 models (\texttt{text-davinci-002} and \texttt{text-davinci-003}) are not included in this analysis.}
\label{tab:experiments:results:correlations}
\end{table}

All tested models always prefer the incorrect output resulting in zero classification accuracy, the log-likelihood of the incorrect output is always significantly higher than the uniform baseline, but it varies with the model.
Specifically:
\begin{itemizesquish}{-0.3em}{0.5em}

    \item The Meta AI OPT and OpenAI text-based GPT-3 families exhibit strong inverse scaling, with the larger models more strongly preferring the incorrect output. The trend is monotonic for the "First generation" GPT-3 family, and somewhat non-monotonic for the OPT and InstructGPT families. The InstructGPT models perform worse than the base GPT-3 models.
    \item The Salesforce CodeGen models exhibit mostly flat scaling. The ``mono'' models which are further fine-tuned on Python-only data perform worse than the "multi" models they are based on.
    \item The OpenAI Codex models are the only models that seem to show positive scaling (which may be spurious since they are only two data points). However, the two GPT-3.5 models (\texttt{text-davinci-002} and \texttt{text-davinci-003}, shown in the figures as red crosses) that further fine-tune \texttt{code-davinci-002} on English demonstrations, lose their edge and end up performing worse than the base GPT-3 model of the same size (\texttt{davinci}).
    \item Google FLAN-T5 shows an unclear, oscillating scaling trend, with large error bars at each point.
\end{itemizesquish}

We report numerical correlation results between model size and mean loss\footnote{in the log-log scale, which for Pearson's correlation measures the adherence to the (inverse of) power law scaling as described by \citet{Scaling_laws_LM}.} in Table~\ref{tab:experiments:results:correlations}.
Due to the small number of model sizes per family, some of the p-values are quite high, but the numerical results are consistent with the qualitative analysis.

Overall, our analysis shows that autoregressive text-based LLMs (even when previously pretrained on code-based models) exhibit inverse scaling on our task, while the code-based models exhibit flat scaling which might possibly transition to positive scaling at the largest tested size, but  fail to substantially improve over the text-based models.

\subsection{Chat LLMs Accuracy}
\label{SEC:EXPERIMENTS:CHATLLMS}

We perform additional experiments on chat LLMs by OpenAI and Anthropic, whose APIs became recently available.
These models constrain both the input text and the generated output to take the form of a dialogue between the user and  the "assistant" (the model itself).
Notably, the APIs of these models do not report log-probabilities, hence they cannot be used to score arbitrary texts.
This prevents us from using the same experimental protocol of the other experiments.
We instead reformulate the task as binary classification where the model is presented with both the correct and incorrect forms of the same program in the same user message and is asked to select the correct one.
We describe the models and the prompt templates in Appendix \ref{SEC:EXPERIMENTS:CHATLLMS_APPENDIX}.

\paragraph{Results}

\begin{figure}[ht]
\centering
\includegraphics[width=0.47\textwidth]{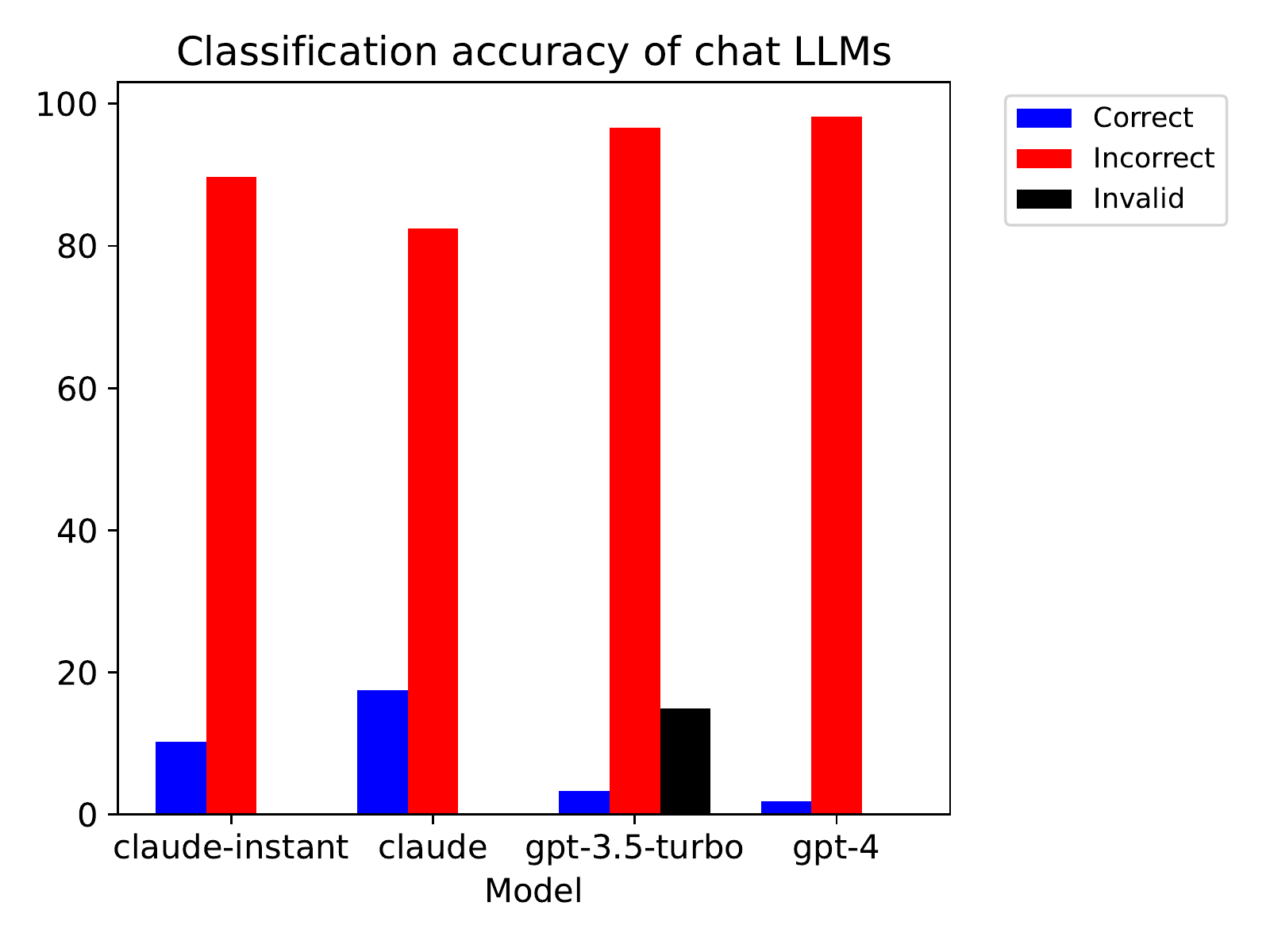}
\caption{Classification accuracy on chat LLMs. For each model, we report the percentage of correct, incorrect and invalid outputs.}
 \label{fig:experiments:results:non_builtins}
\end{figure}

We report the results in Figure \ref{fig:experiments:results:chatllms}.
All the models strongly prefer the incorrect programs, although the classification accuracy is non-zero.
This may not be necessarily comparable to the zero classification accuracy of the previous experiments, due to the different experimental protocol.
The Anthropic models (\texttt{claude-instant} and \texttt{claude}) show better accuracy (10-18\%) with positive scaling and never produce invalid outputs.
The OpenAI models (\texttt{gpt-3.5-turbo} and \texttt{gpt-4}) show low accuracy (< 4\%) with flat or inverse scaling and occasionally produce invalid outputs.

\subsection{Qualitative Experiments}
\label{SEC:EXPERIMENTS:QUALITATIVE}

We perform a small number of manual two-shot experiments on GPT-3.5.
We also carry out manual experiments on  OpenAI ChatGPT-3.5\footnote{\url{https://openai.com/blog/chatgpt/}} and GPT-4 models, where we interact with the models in multiple rounds of dialogue, trying to hint the correct solution.
The models are still unable to provide the correct continuations. See Appendices~\ref{sec:appendix:fewshotexample}--\ref{sec:appendix:chatgpt4example}.

\section{Related work}
\label{SEC:RELATED}

\ignore{LLMs are becoming increasingly popular for a variety of tasks.
This begs the question of how reliable these models are, particularly in settings where providing incorrect answers may cause substantial harm to the users, such as providing medical advice or being making automated decisions in institutional settings that can affect the quality of life of people.}
%These models are used due to their capability of encoding significant amounts of data, allowing them to generate comprehensive and sophisticated results that are factually in correct. If the capabilities of these models were to extrapolate they can cause unintended risks
Recent work sought to characterize the quality of LLMs on a variety of tasks: 
BIG-bench \cite{LLM_x_risk_extrap} is a large collaboration which resulted in a suite of hard, disparate tasks which were used to evaluate various LLMs.
The study found that scaling can be slower and less smooth than expected by naive scaling laws, and social biases sometimes show inverse scaling, also observed by \citet{ParrishBias2021}.
\citet{true_fs_llms} investigated the effect of example selection in few-shot learning for LLMs, finding that previous studies generally overestimated model quality due to methodological issues.
\citet{truthful_QA} attempted to measure the \textit{truthfulness} of the answer provided by LLMs on tasks involving real-world knowledge, finding that while larger models tend to provide more informative answers, they also tend to be less truthful. However, this effect might be confounded due to the dataset design to specifically be adversarial for the largest model being evaluated \cite{kilcher2021}. \newcite{li2023bert} showed that similar to our case, mathematical article processing is sensitive to semi-invariant symbol replacements.
\citet{JiHallucination2022} provide a broad survey about hallucination (generation of fluent yet incorrect information) by natural language generation models.

\section{Conclusions}
We explored the ability of large language models to predict the correct continuations of fragments of Python programs in scenarios where the correct continuations are statistically uncommon due to the redefinition of identifiers caused by a statement that we included in the prompt.
Not only all the tested models fail at this task, but some model families even display \emph{inverse scaling}: they become worse, rather than better, with increasing model size.
These results suggest that LLMs rely on ``shortcut learning'', i.e., weak, unstable, mostly lexical correlations in the data, rather than an understanding of the semantics of the data (in this case, Python code) at a deep level. We believe that our results are important both for a better scientific understanding of the capabilities of LLMs and for their practical relevance as a core technology for automated code generation tools.
Future work could investigate scaling effects at larger model sizes, as well as on other programming languages.

\section*{Limitations}
Our approach has the following limitations:
\begin{enumerate}
    \item It only considers swaps of pairs of functions at the top-level scope, which is a small set of all the quasi-invariances of the Python programming language.
    \item It only considers code generation in top-level functions, hence it does not evaluate class methods.
    \item It relies on a syntactic substitution to generate "correct" gold truth outputs, which may fail if the swapped functions are called by a string expression through \texttt{eval} or or queried by their string names using the reflection facilities.
    \item In our experiments, we can evaluate only a small number of model sizes per family, since these are the only ones available, therefore the p-values of the correlation with the loss analysis are high.
    \item The independent reproducibility of the experiments on closed-source models is predicated on the continued availability of a publicly-accessible API. At the time of writing, our experiments on the OpenAI "Codex" models are no longer reproducible without support from OpenAI.
\end{enumerate}
Items 1 and 2 can be in principle treated by considering more complex code transformations, which we leave for future work.
Item 3 is harder to tackle in the general case because of undecidability issues.
Item 4 could be addressed by reproducing our experiments on a model family that encompasses more model sizes, should it become available for public experimentation.
Item 5 is an unavoidable consequence of using closed-source models.

%First, the approach it only considers swaps of pairs of builtin functions in top-level functions, so it cannot be used to evaluate functions with more complicated structures.
%Second, the exact behaviour of the function may not be accurately captured since the substitution is randomly chosen, and the model is not aware of the context of the substitution. 
%Thirdly, the approach only works with top-level functions, and not with functions nested inside other functions. Finally, the nature of GitHub code means that the data generated may be  of variable quality and not all code snippets may be valid Python code. 
%ACL 2023 requires all submissions to have a section titled ``Limitations'', for discussing the limitations of the paper as a complement to the discussion of strengths in the main text. This section should occur after the conclusion, but before the references. It will not count towards the page limit.
%The discussion of limitations is mandatory. Papers without a limitation section will be desk-rejected without review.

% While we are open to different types of limitations, just mentioning that a set of results have been shown for English only probably does not reflect what we expect. 
% Mentioning that the method works mostly for languages with limited morphology, like English, is a much better alternative.
% In addition, limitations such as low scalability to long text, the requirement of large GPU resources, or other things that inspire crucial further investigation are welcome.

\section*{Ethics Statement}
We do not perform experiments on human subjects.
Our work involves generating a dataset of public data scraped from the GitHub and evaluating it on multiple large language models.
We release our dataset and the code used to generate it.
We filtered our dataset to make sure that all the data that we used has been relesed under the CC-BY-4.0 license, which in our understanding allows for re-releasing, however our filtering procedure is heuristic which implies that there is the possibility that some of the included data may be in violation of its license.
In order to mitigate this hazard, we provide a clearly documented takedown option on the repository on which we will host this data, enabling people to claim copyright and ask for removal of their data.

\section*{Acknowledgements}
We thank the reviewers for their helpful comments.
We thank the Inverse Scaling Prize competition organisers \cite{mckenzie2022inverse} for organising the challenge and donating part of the OpenAI API credits that were used in our experiments.
We are grateful to Apart Research\footnote{\url{https://apartresearch.com/}} for their donation that supported the purchase of additional OpenAI API credits and provided personal financial support to Antonio Valerio Miceli-Barone.
This work was supported by the UKRI Research Node on Trustworthy Autonomous Systems Governance and Regulation (grant EP/V026607/1) which provided funding for Antonio Valerio Miceli-Barone.
The experiments in this work on open source LLMs were supported by a compute grant (UKRI HPC) from the Baskerville service at the University of Birmingham.
%\newpage
\bibliography{anthology,custom}
\bibliographystyle{acl_natbib}
%\newpage

\appendix

\section{Models}
\label{SEC:EXPERIMENTS:MODELS}

%\subsubsection{GPT-3}
\label{SEC:EXPERIMENTS:MODELS:GPT3}

\paragraph{GPT-3} LLMs in the OpenAI GPT-3 family, available in different generations:\footnote{The publicly-available OpenAI models may differ from those described in the papers. Refer to \url{https://beta.openai.com/docs/models/gpt-3} and \url{https://beta.openai.com/docs/model-index-for-researchers} for a detailed description.}
\begin{itemizesquish}{-0.3em}{0.5em}

    \item ``First generation'' GPT-3 \citep{openai2020gpt3}, trained with an unsupervised maximum likelihood estimate next-token prediction objective on raw, byte-pair-encoding tokenized \citep{sennrich-etal-2016-neural} text crawled from the web.
    We use the four models available on the public API: \texttt{ada} (0.35B parameters), \texttt{babbage} (1.3B), \texttt{curie} (6.7B) and \texttt{davinci} (175B).
    \item ``Second generation'' InstructGPT \citep{openai2022instructgpt}, fine-tuned on human-written demonstrations and human-vetted samples (OpenAI ``FeedME'' approach).
    \texttt{text-ada-001} (0.35B), \texttt{text-babbage-001} (1.3B), \texttt{text-curie-001} (6.7B) and \texttt{text-davinci-001} (175B).
    \item ``Third generation'' GPT-3.5. Two ``Codex'' models trained on code, similar to \citet{openai2021codex}: \texttt{code-cushman-001} (12B) and \texttt{code-davinci-002} (175B), and two models based on \texttt{code-davinci-002} and further fine-tuned on human demonstrations with FeedME and PPO \citep{openai2017ppo}, respectively: \texttt{text-davinci-002} and \texttt{text-davinci-003} (both 175B). Unfortunately, at the time of writing, the Codex models are no longer available on the OpenAI API.
\end{itemizesquish}

Our experiments on the OpenAI models were performed with their public API, at a cost of approximately 90 USD.

%\subsubsection{CodeGen}
\label{SEC:EXPERIMENTS:MODELS:CODEGEN}
\paragraph{CodeGen} Salesforce CodeGen models\footnote{From Hugging Face: \url{https://huggingface.co/docs/transformers/model_doc/codegen}} \citep{Nijkamp2022codegen}.
CodeGen is available in two families of auto-regressive LMs:
\begin{itemizesquish}{-0.3em}{0.5em}
    \item \texttt{codegen-X-multi}: first pretrained on the Pile \citep{Gao2021thePile}, an English text corpus, then fine-tuned on a corpus of multiple programming languages. We use the four available model sizes: 0.35B, 2B, 6B and 16B.
    \item \texttt{codegen-X-mono}: based on the ``multi'' models of corresponding size and further fine-tuned on Python data: 0.35B, 2B, 6B and 16B.
\end{itemizesquish}

%\subsubsection{OPT}
\label{SEC:EXPERIMENTS:MODELS:OPT}
\paragraph{OPT} Meta AI OPT models\footnote{From Hugging Face: \url{https://huggingface.co/docs/transformers/model_doc/opt}} \citep{Zhang2022opt}, a family of auto-regressive LMs predominantly trained  on English text.
We use the six available model sizes: 0.125B, 0.35B, 1.3B, 2.7B, 6.7B and 13B.

%\subsubsection{Flan-T5}
\label{SEC:EXPERIMENTS:MODELS:FLANT5}
\paragraph{FLAN-T5} Google FLAN-T5 sequence-to-sequence models \citep{Chung2022FLAN}, obtained by fine-tuning the T5 models on a large number of tasks.
The T5 models \citep{Raffel2019T5} are themselves pretrained on a combination of unsupervised language modeling (formulated as denoising autoencoding) and multiple supervised tasks.
We evaluate each example in our dataset by presenting the prompt (swap statement, function declaration and docstring) as an input to the encoder and ``good'' and ``bad'' classes as alternative inputs to the decoder, for which the model computes the likelihoods.
We consider the following models:\footnote{From Hugging Face: \url{https://huggingface.co/docs/transformers/model_doc/flan-t5}} \texttt{flan-t5-small} (0.08B), \texttt{flan-t5-base} (0.25B), \texttt{flan-t5-large} (0.78B), \texttt{flan-t5-xl} (3B) and \texttt{flan-t5-xxl} (11B).
\\

Our experiments on the CodeGen, OPT and FLAN-T5 models were performed on the Baskerville Tier 2 HPC platform.

\section{Experiment on Non-builtin Functions}
\label{SEC:EXPERIMENTS:NONBUILTINS}

\begin{figure}[ht]
\centering
\includegraphics[width=0.47\textwidth]{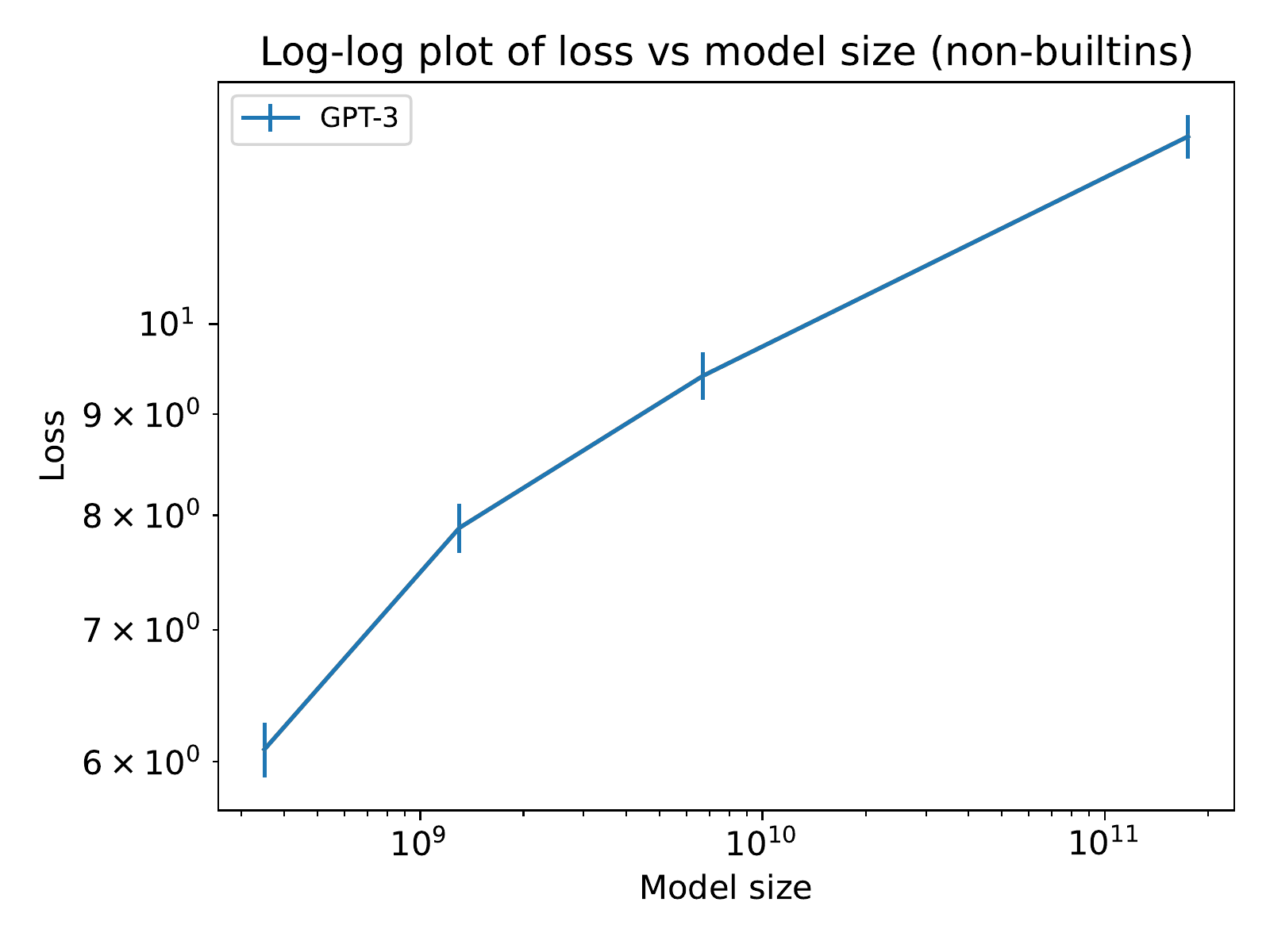}
\caption{Swap of non-builtin functions. Classification loss over model size. First generation GPT-3 models.}
 \label{fig:experiments:results:chatllms}
\end{figure}

We report an additional variant of our main quantitative experiment, evaluating the effect of swapping non-builtin top-level functions.
These functions are defined either within the repositories we sample from, or in external libraries.
Since these function are less standardized compared to the Python builtins, the question arises of whether the models would display a similarly strong preference on their use, ignoring the swap statements. Due to time and cost constraints, we only run this experiment on the first generation GPT-3 models: \texttt{ada}, \texttt{babbage}, \texttt{curie} and \texttt{davinci}.

\paragraph{Results} We report the results in Figure \ref{fig:experiments:results:non_builtins}.
The results are qualitatively equivalent to those of the experiments on the builtin functions: all the models prefer the incorrect but common usage of the swapped functions to the correct but out-of-distribution one, demonstrating inverse scaling.
The overall magnitude of the losses however is smaller, indicating that the preference for the incorrect usage is not as strong as in the builtin function experiment.

\newpage
\onecolumn
\section{Quantitative Chat LLM Experiments}
\label{SEC:EXPERIMENTS:CHATLLMS_APPENDIX}

The most recent LLMs by OpenAI and Anthropic available on their respective public APIs are Chat LLMs.
These models are constrained such that their input must be a sequence of messages of a dialogue between the user and an "assistant" representing the model itself\footnote{an additional "system" role can be used in the messages to specify the general task the LLM is expected to perform.}.
The model then responds with a single message written in the "assistant" role.
Notably, these systems don't report log-probabilty scores, which prevents us from computing a cross-entropy loss for the correct continuation as we did in our main set of experiments, therefore we reformulated the task as a binary choice between two options provided in the same "user" message.

\subsection*{Prompt}

For each example present both the correct program and the incorrect program and ask the model to determine which one is more likely to be correct.
We use the following prompt template:

\begin{tcolorbox}[coltitle=black,colbacktitle=emerald,title=System]
\begin{verbatim}
You are a helpful assistant.
\end{verbatim}
\end{tcolorbox}

\begin{tcolorbox}[coltitle=black,colbacktitle=babyblue,title=User]
\begin{verbatim}
Consider the following Python programs:

Program 1:

{program_1}

Program 2:

{program_2}

Which program is more likely to be correct? Write only the number of the program 
and nothing else.
\end{verbatim}
\end{tcolorbox}

where \verb|{program_1}| and \verb|{program_2}| are replaced by the correct program and the incorrect program, each consisting of the the swap statement, the function declaration and the docstring, followed by either the correct or incorrect function body.

We submit this prompt to the model, setting sampling temperature to $0.0$, then we parse the "assistant" reply, extracting the first integer to evaluate the output.
If don't find any integer or the integer is neither 1 nor 2, we consider the response invalid.
In order to avoid any bias dependent on the order in which the programs are presented, each example is evaluated twice, once with the correct program in the \verb|{program_1}| position, and once in the \verb|{program_2}| position.

\subsection*{Models}

\begin{itemize}
    \item OpenAI \texttt{gpt-3.5-turbo} version 0301 powers the free version of ChatGPT and is believed to be a compressed version of \texttt{text-davinci-003} with additional instruction fine-tuning. \texttt{gpt-4} version 0314 \citep{openai2023gpt4} is the most recent and highest quality OpenAI model available at the time of writing. The API cost of running our experiments was about 40 USD.
    \item Anthropic \texttt{claude-instant} version 1.1 and \texttt{claude} version 1.3 are chat LLMs trained on human feedback\footnote{\url{https://console.anthropic.com/docs}}. \texttt{claude-instant} is described as "low latency" while \texttt{claude} is described as "superior reasoning".
\end{itemize}

The parameter number and training set sizes of these models are not publicly known, however it is plausible that \texttt{gpt-4} is larger than \texttt{gpt-3.5-turbo} and \texttt{claude} is larger than \texttt{claude-instant}.

We report the results of our experiments in \S\ref{SEC:EXPERIMENTS:CHATLLMS}.

\section{Dataset Example}
\label{sec:appendix:datasetexample}
We present one full example from out dataset.
%\ref{fig:appendix:datasetexample}).

%\newpage
%\onecolumn

%\begin{figure*}[t]
\begin{tcolorbox}[coltitle=black,colbacktitle=babyblue,title=Head]
\begin{minted}[autogobble,numberblanklines=false,frame=lines]{python}
len, open = open, len
def importfile(path):
    """Import a Python source file or compiled file given its path.""""
\end{minted}
\end{tcolorbox}
%\end{center}

\begin{tcolorbox}[coltitle=black,colbacktitle=bananayellow,title=Incorrect continuation]
%\begin{center}
\begin{minted}[numberblanklines=false,frame=lines]{python}
    from importlib.util import MAGIC_NUMBER
    with open(path, 'rb') as ifp:
        is_bytecode = (MAGIC_NUMBER == ifp.read(len(MAGIC_NUMBER)))
    filename = os.path.basename(path)
    (name, ext) = os.path.splitext(filename)
    if is_bytecode:
        loader = importlib._bootstrap_external.SourcelessFileLoader(name, path)
    else:
        loader = importlib._bootstrap_external.SourceFileLoader(name, path)
    spec = importlib.util.spec_from_file_location(name, path, loader=loader)
    try:
        return importlib._bootstrap._load(spec)
    except ImportError:
        raise Exception(path, sys.exc_info())
\end{minted}
\end{tcolorbox}
%\end{center}

\begin{tcolorbox}[coltitle=black,colbacktitle=bananayellow,title=Correct continuation]
%\begin{center}
\begin{minted}[numberblanklines=false,frame=lines]{python}
    from importlib.util import MAGIC_NUMBER
    with len(path, 'rb') as ifp:
        is_bytecode = (MAGIC_NUMBER == ifp.read(open(MAGIC_NUMBER)))
    filename = os.path.basename(path)
    (name, ext) = os.path.splitext(filename)
    if is_bytecode:
        loader = importlib._bootstrap_external.SourcelessFileLoader(name, path)
    else:
        loader = importlib._bootstrap_external.SourceFileLoader(name, path)
    spec = importlib.util.spec_from_file_location(name, path, loader=loader)
    try:
        return importlib._bootstrap._load(spec)
    except ImportError:
        raise Exception(path, sys.exc_info())
\end{minted}
\end{tcolorbox}
%\label{fig:appendix:datasetexample}
%\end{figure*}
The statement at the beginning of the prompt swaps the identifiers for the builtin \texttt{len} and \texttt{open} functions.
The incorrect continuation still uses the \texttt{open} function to open the file and \texttt{len} to compute the length of the byte array \texttt{MAGIC\_NUMBER}, in the correct continuation instead, \texttt{open} and \texttt{len} are swapped.

\section{Example of Few-shot Interaction with GPT-3}
\label{sec:appendix:fewshotexample}

We report an example of a manual few-shot interaction with GPT-3.5 \texttt{text-davinci-003} where we provide two correct examples in the prompt and we ask the model to generate a continuation on a third example.
This task should be easier than the ones we consider in our dataset because the two in-context examples should cue the model about the need to swap the builtin function identifiers in the function bodies.
Nevertheless, the model fails to swap the builtins in the generated continuation.
Running the example multiple times in order to account for sampling randomness results in slightly different but functionally equivalent continuations, all containing the same mistake.
Switching to \texttt{code-davinci-002} also does not help.
%\todo{can you say a bit more, does it fail or succeed, etc.?}

%\begin{figure*}[t]

\begin{tcolorbox}[coltitle=black,colbacktitle=babyblue,title=Prompt]
\begin{minted}[autogobble,numberblanklines=false,frame=lines]{python}
len, print = print, len
def print_len(x):
  "Print the length of x"
  len(print(x))

def foo(x):
  "Print 'foo' if the length of x is even, otherwise print 'bar' "
  if print(x) % 2 == 0:
    len('foo')
  else:
    len('bar')

any, all = all, any
def bar(x, y)
  "Return true if all the elements of x are \
  positive numbers and any element of y is a negative number"
\end{minted}
\end{tcolorbox}

\begin{tcolorbox}[coltitle=black,colbacktitle=bananayellow,title=Generated continuation]
\begin{minted}[numberblanklines=false,frame=lines]{python}
   return all(x > 0) and any(y < 0)
\end{minted}
%\label{fig:appendixfewshotexample}
\end{tcolorbox}

\section{ChatGPT-3.5 Example}
\label{sec:appendix:chatgptexample}

We present an interaction with OpenAI ChatGPT\footnote{While we run this experiment on the ChatGPT web interface \url{https://chat.openai.com/}, this model is presumed to be equivalent to \texttt{gpt-3.5-turbo-0301} available on the OpenAI API and Playground web interface}.
The model not only fails to generate the correct code completion on the first try, but still fails even when the error is pointed out in multiple rounds of dialogue.
It fails even when prompted with the \texttt{Let's think step by step} phrase \citet{Kojima2022ZeroShot} found to improve zero-shot "reasoning" abilities.

%\begin{figure*}[t]
%\twocolumn %% we cannot have this command inside a figure* -- it causes the compiler to hang. -- Shay
\begin{tcolorbox}[coltitle=black,colbacktitle=babyblue,title=User]
Complete the following Python program:

\begin{minted}[autogobble,numberblanklines=false,frame=lines]{python}
len, print = print, len
def print_len(x):
  "Print the length of x"
  # complete this function
\end{minted}  
\end{tcolorbox}

\begin{tcolorbox}[coltitle=black,colbacktitle=bananayellow,title=Generated response]
\begin{minted}[autogobble,numberblanklines=false,frame=lines]{python}
def print_len(x):
  "Print the length of x"
  print(len(x))}
\end{minted}
\end{tcolorbox}

\begin{tcolorbox}[coltitle=black,colbacktitle=babyblue,title=User]
what does the first line of the program that I've written do?
\end{tcolorbox}

\begin{tcolorbox}[coltitle=black,colbacktitle=bananayellow,title=Generated response]
The first line of the program is swapping the values of the variables len and print. This means that len will now have the value of the print function and print will have the value of the len function. This means that if you call len in the program, it will actually run the print function, and if you call print, it will run the len function. This is known as "aliasing" or "renaming" variables.
\end{tcolorbox}

\begin{tcolorbox}[coltitle=black,colbacktitle=babyblue,title=User]
so, was you program completion correct?
\end{tcolorbox}

\begin{tcolorbox}[coltitle=black,colbacktitle=bananayellow,title=Generated response]
Yes, my completion of the program is correct. The function print\_len simply calls the print function (which is actually the len function due to the variable swapping in the first line of the program) and passes in the argument x. This will print the length of x.
\end{tcolorbox}

\begin{tcolorbox}[coltitle=black,colbacktitle=babyblue,title=User]
should it be print(len(x)) or len(print(x)) ?
\end{tcolorbox}

\begin{tcolorbox}[coltitle=black,colbacktitle=bananayellow,title=Generated response]
It should be print(len(x)). The print function (which is actually the len function due to the variable swapping in the first line of the program) is called and passed the argument x. The len function is then called on x and the result of this function is passed as an argument to the print function, which prints the result.If it were len(print(x)), the print function (which is actually the len function) would be called with x as the argument, and then the result of this function (which would be the length of x) would be passed as an argument to the len function, which would simply return the length of the result (which would always be 1 since the result is a number). This would not produce the desired output.
\end{tcolorbox}

\begin{tcolorbox}[coltitle=black,colbacktitle=babyblue,title=User]
Are you sure about this reasoning? Let's think step by step
\end{tcolorbox}

\begin{tcolorbox}[coltitle=black,colbacktitle=bananayellow,title=Generated response]
Yes, I am sure about this reasoning. Here is a step-by-step breakdown of what happens when the print\_len function is called: The print\_len function is called with an argument x. The print function is called with the argument len(x). The len function is called on x and returns the length of x. The print function (which is actually the len function due to the variable swapping in the first line of the program) is called with the result of the len function (the length of x) as the argument. The print function prints the result (the length of x). If it were len(print(x)), the following would happen:The print\_len function is called with an argument x. The print function (which is actually the len function due to the variable swapping in the first line of the program) is called with the argument x. The print function (which is actually the len function) returns the length of x. The len function is called with the result of the print function (the length of x) as the argument. The len function returns the length of the result (which would always be 1 since the result is a number). This does not produce the desired output of printing the length of x.
\end{tcolorbox}

\begin{tcolorbox}[coltitle=black,colbacktitle=babyblue,title=User]
Point 3. of your explanation of print(len(x)) is incorrect because print and len are swapped
\end{tcolorbox}

\begin{tcolorbox}[coltitle=black,colbacktitle=bananayellow,title=Generated response]
You are correct, point 3 of my explanation for print(len(x)) is incorrect because print and len are swapped. Here is the corrected step-by-step breakdown: The print\_len function is called with an argument x. The len function (which is actually the print function due to the variable swapping in the first line of the program) is called with the argument x. The print function (which is actually the len function due to the variable swapping in the first line of the program) returns the length of x. The print function (which is actually the len function) is called with the result of the len function (the length of x) as the argument. The print function (which is actually the len function) prints the result (the length of x).
\end{tcolorbox}

\begin{tcolorbox}[coltitle=black,colbacktitle=babyblue,title=User]
in point 5, how can print , which is actually len, print anything?
\end{tcolorbox}

\begin{tcolorbox}[coltitle=black,colbacktitle=bananayellow,title=Generated response]
You are correct that in point 5, it is incorrect to say that the print function (which is actually the len function) is printing anything. The correct statement would be that the len function (which is actually the print function due to the variable swapping in the first line of the program) is printing the result (the length of x). I apologize for the confusion. Here is the corrected step-by-step breakdown: The print\_len function is called with an argument x. The len function (which is actually the print function due to the variable swapping in the first line of the program) is called with the argument x. The print function (which is actually the len function due to the variable swapping in the first line of the program) returns the length of x. The print function (which is actually the len function) is called with the result of the len function (the length of x) as the argument. The len function (which is actually the print function) prints the result (the length of x).
\end{tcolorbox}

\begin{tcolorbox}[coltitle=black,colbacktitle=babyblue,title=User]
good, so the correct program is
\end{tcolorbox}

\begin{tcolorbox}[coltitle=black,colbacktitle=bananayellow,title=Generated response]
Yes, the correct program is:

\begin{minted}[autogobble,numberblanklines=false,frame=lines]{python}
len, print = print, len
def print_len(x):
  "Print the length of x"
  print(len(x))
\end{minted}

This program will correctly swap the values of the len and print variables, and then call the print function (which is actually the len function due to the variable swapping) with the argument x. The len function will be called on x and the result (the length of x) will be passed as an argument to the print function, which will then print the result.
\end{tcolorbox}

\begin{tcolorbox}[coltitle=black,colbacktitle=babyblue,title=User]
Wrong.
\end{tcolorbox}

\begin{tcolorbox}[coltitle=black,colbacktitle=bananayellow,title=Generated response]
I apologize if my previous explanations were incorrect. Please let me know how I can better assist you in understanding the program.
\end{tcolorbox}

\section{GPT-4}
\label{sec:appendix:chatgpt4example}
We present an interaction with OpenAI GPT-4 version 0314.\footnote{\url{https://platform.openai.com/playground?mode=chat&model=gpt-4-0314}}. 
We use a temperature of 1.0. 
GPT-4 is the most recent and most generally accurate OpenAI LLM available at the time of writing, nevertheless it fails to generate the correct program even when cued by the user. We note that our information about the inner-workings of GPT-4 are more limited than with the previous versions of the GPT model family.

\begin{tcolorbox}[coltitle=black,colbacktitle=emerald,title=System]
You are an helpful assistant.
\end{tcolorbox}

\begin{tcolorbox}[coltitle=black,colbacktitle=babyblue,title=User]
Complete the following Python program:
\begin{minted}[autogobble,numberblanklines=false,frame=lines]{python}
len, print = print, len

def print_len(x):

  "Print the length of x"

  # complete this function
\end{minted}
\end{tcolorbox}

\begin{tcolorbox}[coltitle=black,colbacktitle=bananayellow,title=Generated response]
Here's the complete Python program:

\begin{minted}[autogobble,numberblanklines=false,frame=lines]{python}

len, print = print, len

def print_len(x):

    "Print the length of x"

    print(len(x))

# Example usage

test_string = "Hello, World!"

print_len(test_string)
\end{minted}
This program swaps the `print()` and `len()` functions, and then defines a function `print\_len()` that takes a single argument `x` and prints its length using the swapped `print()` and `len()` functions.
\end{tcolorbox}

\begin{tcolorbox}[coltitle=black,colbacktitle=babyblue,title=User]

what does the first line of the program that I’ve written do?
\end{tcolorbox}
\begin{tcolorbox}[coltitle=black,colbacktitle=bananayellow,title=Generated response]
The first line of the program you've written swaps the `print` and `len` functions:
\begin{minted}[autogobble,numberblanklines=false,frame=lines]{python}

len, print = print, len

\end{minted}
Here, you are creating a tuple with `(print, len)` and then unpacking that tuple into the variables `len` and `print`. As a result, the `print` variable now holds the built-in `len` function, and the `len` variable holds the built-in `print` function.

After this line, whenever you use `len` in your code, it will call the 'print' function, and when you use `print`, it will call the 'len' function. This can be confusing and is generally not recommended for writing clear, maintainable code, but it seems that the intention here is to create a playful or unconventional challenge.
\end{tcolorbox}

\begin{tcolorbox}
[coltitle=black,colbacktitle=babyblue,title=User]

so, was you program completion correct?
\end{tcolorbox}

\begin{tcolorbox}

Apologies for the confusion in my initial response. Since the `print()` and `len()` functions were swapped in the first line, my initial completion was incorrect. Here's the corrected completion:
\begin{minted}
[autogobble,numberblanklines=false,frame=lines]{python}    
len, print = print, len

def print_len(x):
    "Print the length of x"
    len(x)  # since len now refers to print function
  
# Example usage
test_string = "Hello, World!"
print_len(test_string)
\end{minted}

Now, in the `print\_len()` function, the `len` function is used to print the length of the input `x` because, after the first line, `len` actually refers to the original `print` function.
\end{tcolorbox}

\begin{tcolorbox}
[coltitle=black,colbacktitle=babyblue,title=User]
is len(x) correct?
\end{tcolorbox}

\begin{tcolorbox}
[coltitle=black,colbacktitle=bananayellow,title=Generated response]

Apologies again, I realize I made a mistake in my previous response. Since the functions are swapped, you should use the `print` function, which now refers to the original `len` function, to get the length of the input `x`. Here's the corrected completion:

\begin{minted}[autogobble,numberblanklines=false,frame=lines]{python}
len, print = print, len

def print_len(x):
    "Print the length of x"
    len(print(x))  # since print now refers to len function

# Example usage
test_string = "Hello, World!"
print_len(test_string)
\end{minted}
In the `print\_len()` function, we now use `len(print(x))`, which calculates the length of `x` (because `print` now refers to the original `len` function) and then prints it using the swapped `len` function (which now refers to the original `print` function).
\end{tcolorbox}

\section{Program Equivariances and Invariances}
\label{sec:appendix:equivinv}

In this section we provide the formal definition of program equivariances and invariances, and specifically of $\alpha$-equivalence, which the identifier swaps in Python lead to.
The definition relies on the notion of syntactic transformations %that do not change the meaning of the program. In the case of identifier swaps, these transformations 
which can be formalized as the algebraic structure of a group.

\paragraph{Group action}
Let $G$ be a group with identity element $\epsilon$ and $X$ be a set.
The function $T : G \times X \to X$ is a \textit{(left) group action} of $G$ on $X$ if $\forall x \in X, g \in G, h \in G$
\begin{align*}
    T(\epsilon, x) &= T(x) \\
    T(g \cdot h, x) &= T(g, T(h, x))
\end{align*}
Intuitively, $T$ is a transformation on the elements of $X$ which is parameterized by the elements of group $G$ on the in a way consistent with the group structure, so that the identity element corresponds to the identity transformation and combining the transformation parameters with their own group operation and then applying the result is equivalent to applying them in sequence.

\paragraph{Group equivariance and invariance}
Let $G$ be a group, $X$ and $Y$ be sets.
Let $T : G \times X \to X$ and $S : G \times Y \to Y$ be (left) group actions of $G$ on $X$ and $Y$ respectively.
The function $f : X \to Y$ is \textit{(left) equivariant} w.r.t. group $G$ and $T$ and $S$ if $\forall x \in X, g \in G$
\begin{align*}
    S(g, f(x)) &= f(T(g, x))
\end{align*}
This means that applying the transformation $T$ parameterized by $g$ on $x$ and then evaluating $f$ on it is equivalent to evaluating $f(x)$ first and then transforming the result with $S$ parameterized by $g$. \\
In the special case where $S$ is trivial on the image of $f$, that is $\forall x \in X, g \in G$
\begin{align*}
    S(g, f(x)) &= f(x)
\end{align*}
then $f$ is \textit{(left) invariant} w.r.t. $G$ and $T$, which means that $f$ effectively ignores the transformation $T$ on its inputs.

There has been an interest in recent years in applying these concepts to deep learning, either by measuring the extent to which models spontaneously learn equivariances or by designing model architectures that obey certain equivariances by construction, see \citet{BronsteinGeometric2021} for an extended survey.
 Previous work usually considers equivariances w.r.t. geometrical transformations such as rotations and reflections on data types with a natural physical interpretation, such as images \citep{CohenGECNN2016}, molecules \citep{MusilMolecules2021} or video game grid-world environments \citep{deac2023equivariant}, but the theoretical framework is general enough to encompass many forms of equivariances and data types, including programming code.

\paragraph{$\alpha$-equivalence}
Let $X$ the set of programs (or program fragments) in some language (e.g. Python), let the function $f$ denote their semantics ($f$ can take additional arguments representing the program inputs and environment, and its image is a set of results or sequences of actions that result from the execution of a program).\\
Let $G$ the group of the permutations of all syntactically valid identifier names.
Let $T(g, x)$ the transformation that substitutes the identifiers in program $x$ according to permutation $g$.
If $f$ is invariant w.r.t. $G$ and $T$ then it means that swapping the identifiers inside a program does not affect its execution, a property which is usually called $\alpha$-equivalence in the programming languages literature.

In many programming languages $\alpha$-equivalence may only apply when swapping identifiers in whole programs including the standard library of the language.
Program fragments such as modules, classes or functions (procedures) may not have $\alpha$-equivalence when identifiers defined outside them (e.g. at top-level) are swapped.
In Python however, this effect can be compensated by inserting a swap statement right before the program fragment.
If a permutation $g$ acts non-trivially on top-level identifiers $a_0, a_1, \dots, a_n$, then the tuple assignment statement
\begin{verbatim}
ga_0, ga_1, [..., ga_n] = a_0, a_1, [..., a_n]
\end{verbatim}
will usually make the identifier swap in the program fragment an invariance.
This does not work in all cases because Python programs can access their own source code programmatically and reference identifiers by name from arbitrarily computed strings.
Checking for these conditions is undecidable in the general case, however these are not common programming practices, hence we can ignore them for our purposes.

\end{document}